\documentclass[sigconf]{acmart}

\usepackage{colortbl}

\usepackage{multirow}
\usepackage{array} 
\usepackage{tabularx}
\usepackage{makecell} 

\usepackage{enumitem}
\usepackage{hanging}

\usepackage{algorithm}
\usepackage{algpseudocode}

\usepackage{tcolorbox}
\usepackage{fancyvrb}
\usepackage{listings}

\usepackage{subfigure}
\usepackage{tikz} 

\usepackage{lipsum}

\definecolor{lightblue}{rgb}{0.21, 0.49, 0.74}

\AtBeginDocument{%
  }

\copyrightyear{2026}
\acmYear{2026}
\setcopyright{cc}
\setcctype{by}
\acmConference[CIKM '26] {Proceedings of the 35th ACM International Conference on Information and Knowledge Management}{November 7--11, 2026}{Rome, Italy.}
\acmBooktitle{Proceedings of the 35th ACM International Conference on Information and Knowledge Management (CIKM '26), November 7--11, 2026, Rome, Italy}
\acmISBN{979-8-4007-2539-5/2026/11}
\acmDOI{10.1145/3799682.3840673}

\begin{document}

\title[MKG-CARE: Case-Aware Reasoning with Multimodal Knowledge Graphs for Explainable Medical Image Diagnosis]{MKG-CARE: Case-Aware Reasoning with Multimodal Knowledge Graphs for Explainable Medical Image Diagnosis}


\author{Yiming Xu}
\authornote{Both authors contributed equally to this research.}
\orcid{0009-0003-0918-4037}
\email{xym30@mail.ustc.edu.cn}
\affiliation{%
 \institution{University of Science and Technology of China}
   \city{Hefei}
 \state{Anhui}
 \country{China}
}

\author{Yixuan Liu}
\authornotemark[1]
\orcid{0009-0006-2726-7874}
\email{lyx1022@mail.ustc.edu.cn	}
\affiliation{%
 \institution{University of Science and Technology of China}
   \city{Hefei}
 \state{Anhui}
 \country{China}
}

\author{Yuhang Zhang}
\orcid{0000-0003-2761-8751}
\email{yhzhang@mail.ustc.edu.cn}
\affiliation{%
 \institution{University of Science and Technology of China}
 \city{Hefei}
 \state{Anhui}
 \country{China}
}
\affiliation{%
  \institution{City University of Hong Kong}
  \city{Hong Kong}
  \country{China}
}

\author{Ling Zheng}
\orcid{0009-0008-4819-8710}
\email{lingzheng@mail.ustc.edu.cn}
\affiliation{%
 \institution{University of Science and Technology of China}
 \city{Hefei}
 \state{Anhui}
 \country{China}
}

\author{Yihan Wang}
\orcid{0000-0002-6164-5713}
\email{yihanw@mail.ustc.edu.cn}
\affiliation{%
 \institution{University of Science and Technology of China}
 \city{Hefei}
 \state{Anhui}
 \country{China}
}

\author{Qi Song}
\orcid{0000-0002-1726-7858}
\authornote{Corresponding author.}
\email{qisong09@ustc.edu.cn}
\affiliation{%
 \institution{University of Science and Technology of China}
 \city{Hefei}
 \state{Anhui}
 \country{China}
}

\renewcommand{\shortauthors}{Yiming Xu et al.}

\begin{abstract}
  Medical image diagnosis has achieved significant progress with deep learning, yet existing methods often rely on isolated visual evidence and lack the ability to effectively leverage similar cases and external knowledge. In clinical practice, diagnosis is typically supported by similar historical cases and their associated symptoms. To explicitly model this evidence-based diagnostic process, we propose MKG-CARE, a framework that performs case-aware reasoning using multimodal knowledge graphs for explainable medical image diagnosis. Specifically, we construct a case-aware multimodal knowledge graph as a structured diagnostic memory, where diseases, images, and symptoms are hierarchically organized. Given an input image, MKG-CARE adaptively retrieves similar cases from this memory and extracts their corresponding case-centered subgraphs. We further introduce a knowledge propagation and injection mechanism, where an image-centric Graph Attention Network aggregates heterogeneous semantics within the retrieved case subgraphs, followed by bidirectional cross-modal attention to align and inject the aggregated case knowledge into visual representations. To mitigate retrieval noise, we design a confidence-calibrated decision refinement scheme that estimates each retrieved case’s reliability from prediction confidence and sample similarity, and reweights its contribution to the final prediction for interpretable case-level evidence attribution. Extensive experiments on multiple medical imaging datasets demonstrate consistent improvements over strong baselines, while ablation and qualitative analyses validate the effectiveness and interpretability of our method.
  The code is available at https://github.com/lyxuan1022/MKG-CARE.
\end{abstract}




\maketitle

\section{Introduction}

\begin{figure}[t]         
  \centering
  \includegraphics[
    width=1.0\linewidth,  
    keepaspectratio
  ]{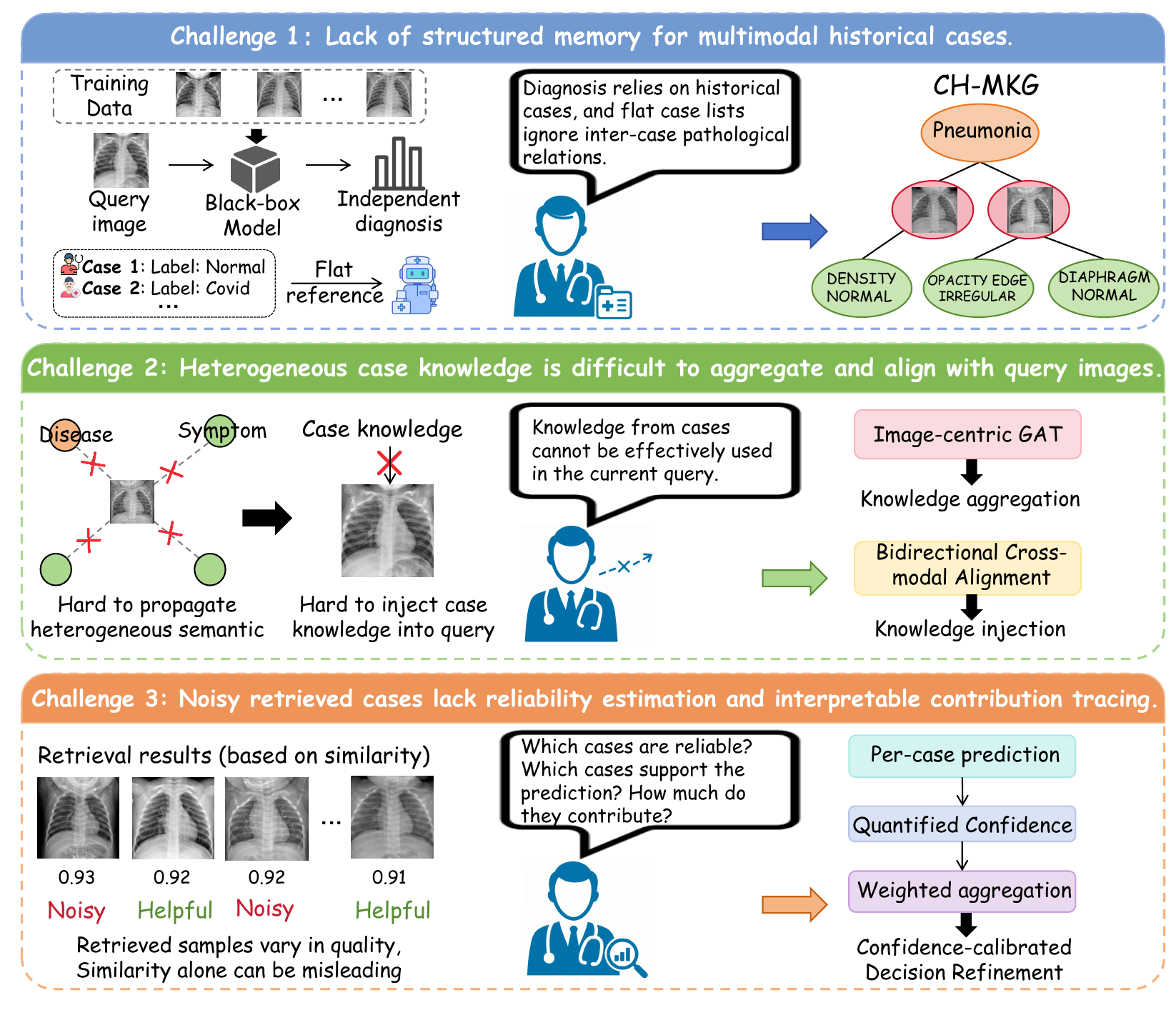}
  
  \caption{Challenges of medical image diagnosis. Each row follows a consistent three-part layout: the left panel illustrates the problem scenario through visual diagrams; the middle panel contains a textual summary of the core issue within a speech bubble; and the right panel presents the corresponding solution pipeline, thereby establishing a clear challenge-to-solution mapping.}
  \label{fig:intro}
\end{figure}

Medical image diagnosis constitutes a core pillar of computer-aided diagnosis (CAD)~\cite{cai2020review,kim2022transfer,wang2026echoagent}, aiming to automatically identify disease-relevant patterns across diverse imaging modalities, including ultrasound~\cite{al2020dataset,snider2022image}, gastrointestinal endoscopy~\cite{ali2024artificial,liu2026endobench}, computed tomography (CT)~\cite{kermany2018large,shastri2022cheximagenet,mao2026ct}, and dermatologic imaging~\cite{joerg2025ai,feng2025ucagents,swapnil2025grpo++}. This process is directly tied to early disease screening and precise clinical intervention. 
While conventional vision-based models~\cite{li2014medical,manzari2023medvit,liu2024vmamba} (e.g., CNNs, ViTs, and Mamba) have achieved significant progress, they predominantly rely on pixel-level feature extraction, often struggling to capture highly context-dependent pathological patterns.
Recently, Large Vision-Language Models (LVLMs), such as HealthGPT~\cite{lin2025healthgpt}, HuatuoGPT~\cite{zhang2023huatuogpt}, GPT-5.1~\cite{openai_gpt_5}, and Gemini~\cite{comanici2025gemini}, have demonstrated immense potential in cross-modal reasoning. However, the development of domain-specific foundation models remains heavily constrained by the scarcity of annotated medical data and prohibitive computational costs. 
Furthermore, such data-driven diagnostic models provide limited evidence traceability and often treat each patient visit as an independent and static event, performing inference solely based on the current imaging input.

To overcome these bottlenecks, incorporating  prior knowledge to guide diagnostic reasoning has emerged as a prominent research trend. The first category of methods~\cite{zhang2025knowledge,huang2025vap} attempts 
to leverage the internal parametric knowledge of Large Language Models (LLMs) to generate multi-dimensional prior descriptions for input images. However, constrained by the inherent “hallucination” issue of LLMs, such automatically generated texts often suffer from factual deviations, leading to the injection of unreliable knowledge.
The second line of research 
introduces external structured knowledge via medical Knowledge Graphs (KGs)~\cite{zheng2026kg, xu2025lgc, chen2025magicdock} to provide rigorous medical concepts, improving the factual consistency of diagnostic reasoning. 
Despite these advances, 
both paradigms largely confine their priors to generated text or abstract medical concepts, while overlooking the reusable diagnostic experience embedded in historical cases.
Real-world clinical diagnosis is inherently an evidence-based and experience-driven process, wherein physicians typically compare similar historical cases to support their final decisions. Motivated by this observation,
a third category focuses on case retrieval, recalling similar historical cases as contextual inputs~\cite{tian2025ecg} to provide intuitive empirical references. As shown in Figure~\ref{fig:intro}, although these attempts mitigate the limitations of purely data-driven models to some extent, establishing a diagnostic framework that truly aligns with clinical reasoning still faces the following three major challenges:

\textbf{Challenge 1: Lack of structured memory for multimodal historical cases.} While referring to historical cases is crucial, most existing methods merely utilize them as training data, implicitly encoding them into model weights. This restricts the inference phase to making independent predictions on isolated query images. Even when recent retrieval-augmented approaches introduce reference samples, they typically treat the retrieved cases as flat, unstructured lists, which severs the lateral connections between cases and overlooks complex pathological correlations. Furthermore, although traditional knowledge graphs attempt to model clinical associations, they are predominantly restricted to text-only medical concepts. This text-only nature not only neglects crucial visual evidence but also makes it exceedingly difficult to cross-modally retrieve relevant structured knowledge directly using visual queries.

\textbf{Challenge 2: Heterogeneous case knowledge is difficult to aggregate and align with query images.} After retrieving similar case subgraphs, effectively bridging the modality gap between continuous visual features and discrete structured semantics emerges as a new technical bottleneck. On the one hand, within a case subgraph, significant representational discrepancies exist between the image entity and its neighboring entities (e.g., symptoms and diseases), hindering the dynamic message passing and semantic aggregation of heterogeneous information. On the other hand, the lack of effective interaction between the current query image and the retrieved cases makes it difficult to deeply integrate the extracted structured case knowledge into the diagnostic reasoning of the current query.

\textbf{Challenge 3: Noisy retrieved cases lack reliability estimation and interpretable contribution tracing.} Although introducing similar cases can supplement clinical evidence, the quality of retrieved samples varies, and noisy or low-relevance cases can easily mislead the model's diagnostic direction. Existing retrieval-augmented methods mostly lack mechanisms to quantify the reliability of auxiliary samples, failing to explicitly measure the specific contribution of each historical case to the final prediction. This implicit fusion approach makes it difficult for physicians to clearly trace whether the current prediction is supported or contradicted by specific historical cases, severely limiting the model's interpretability and trustworthiness in high-risk clinical scenarios.

To address the above challenges, we propose a case knowledge graph-based solution that augments both the data and model levels. \textbf{For Challenge 1}, we construct a case-aware hierarchical multimodal knowledge graph (CH-MKG). Specifically, we select representative images from each disease and integrate multi-dimensional, fine-grained symptom annotations provided by clinical experts to construct a ``disease-image-symptom'' hierarchical graph structure. During the retrieval phase, we design an adaptive contextual retrieval strategy to dynamically define the retrieval boundary by analyzing the similarity gradient between adjacent candidate cases, thereby precisely extracting star-shaped subgraphs with similar images as central entities and associated symptoms and diseases as neighboring nodes. This design effectively filters out interference from low-relevance samples, providing high-quality, structured multimodal prior guidance for the current query image. \textbf{For Challenge 2}, we design a case knowledge propagation and injection module (KPI). First, within the case subgraphs, we employ an image-centric Graph Attention Network~\cite{velivckovic2017graph, jin2026openddi} for explicit message passing, which adaptively aggregates the peripheral symptom and disease semantics into the central image node, generating a structured knowledge representation enriched with clinical context. Subsequently, to achieve deep fusion between the prior knowledge and the current query image, we introduce a dual cross-modal attention mechanism. This mechanism facilitates bidirectional mutual querying and guidance between visual features and knowledge graph, achieving cross-case alignment and deep knowledge injection. \textbf{For Challenge 3}, we introduce a Confidence-calibrated Decision Refinement (CDR) module. This module utilizes a second-pass forward inference mechanism to independently evaluate the retrieved similar cases, dynamically quantifying the reliability of each historical case by synthesizing sample similarity and prediction confidence. Subsequently, it performs an adaptive fusion of the prediction distributions based on these reliability weights. This mechanism not only suppresses the interference of retrieval noise but also explicitly characterizes the specific contribution of each auxiliary case to the final decision.

In summary, our contributions are as follows:
\begin{itemize}[leftmargin=*]
\item We construct a ``disease-image-symptom'' case-aware hierarchical multimodal knowledge graph and combine it with an adaptive retrieval strategy to precisely extract star-shaped local subgraphs as priors. Concurrently, we design a case knowledge propagation and injection module. This module dynamically aggregates heterogeneous semantics within the subgraphs and subsequently achieves deep alignment of cross-case features.
\item We introduce a Confidence-calibrated Decision Refinement module, endowing the model with high interpretability and robustness. Utilizing a second-pass forward inference mechanism, this module dynamically quantifies the reliability of historical cases by synthesizing sample similarity and prediction confidence, followed by adaptive fusion. This not only effectively mitigates retrieval noise but also explicitly traces the contribution of each auxiliary case to the final diagnosis.
\item Extensive quantitative evaluations and qualitative visualization results fully validate the effectiveness of our proposed multimodal knowledge graph and diagnostic framework. The results demonstrate that this method possesses significant advantages in both improving classification performance and enhancing the transparency of the decision-making process.
\end{itemize}

\section{Related Work}

\subsection{Vision-based methods}
A dominant paradigm in medical image classification treats diagnosis as an isolated visual inference problem, where each image is processed independently without external context or historical references. Early approaches are primarily based on convolutional neural networks (CNNs), such as EfficientNet~\cite{tan2021efficientnetv2} and ConvNeXt~\cite{liu2022convnet}, which improve performance through architectural scaling and optimization. Transformer-based models, including Swin Transformer~\cite{liu2021swin}, further enhance global context modeling by capturing long-range dependencies in medical images. More recently, state-space models such as VMamba~\cite{liu2024vmamba} and its medical adaptation MedMamba~\cite{yue2024medmamba} have demonstrated strong efficiency and representation capability for high-resolution medical data. In parallel, large vision-language models (LVLMs)~\cite{zheng2026cat, zheng2026groc, zhang2026medtvt}, including general-purpose models such as GPT~\cite{openai2024chatgpt} and Gemini~\cite{comanici2025gemini}, as well as medical-domain variants such as HealthGPT~\cite{lin2025healthgpt}, HuatuoGPT~\cite{zhang2023huatuogpt}, and MedGemma~\cite{sellergren2025medgemma}, have shown promising cross-modal reasoning ability by leveraging large-scale pretraining. Despite these advances, such methods fundamentally rely on single-image inference, where knowledge is implicitly encoded in model parameters. This paradigm lacks explicit mechanisms to incorporate similar historical cases or structured clinical knowledge, making it difficult to emulate the experience-driven and evidence-based reasoning process of real-world clinical diagnosis.

\subsection{Knowledge-enhanced medical image diagnosis methods}
To overcome the limitations of isolated visual inference, recent studies have incorporated external knowledge and retrieved cases into medical image diagnosis. One line of work leverages textual knowledge; for example, KEM~\cite{zhang2025knowledge} uses LLM-generated symptom descriptions to guide image classification. Another line introduces structured knowledge graphs, such as KG-CMI~\cite{zheng2026kg}, to align lesion features with medical concepts. However, these approaches remain largely text-centric and do not explicitly represent visual cases as graph entities, limiting direct interaction between image evidence and structured knowledge.
Retrieval-based methods further exploit similar historical cases as auxiliary evidence. ECG-Doctor ~\cite{tian2025ecg} retrieves clinically similar samples for prediction, while RareAgents~\cite{chen2026rareagents} extends case retrieval to complex clinical reasoning using phenotypic cases and longitudinal records. Nevertheless, retrieved cases are typically treated as flat contextual references, without explicitly modeling their structured relationships or tracing their individual contributions to the final decision.

\section{Method}

\begin{figure*}[t]
  \centering
  \includegraphics[
    width=\textwidth,
    height=11cm,
    keepaspectratio
  ]{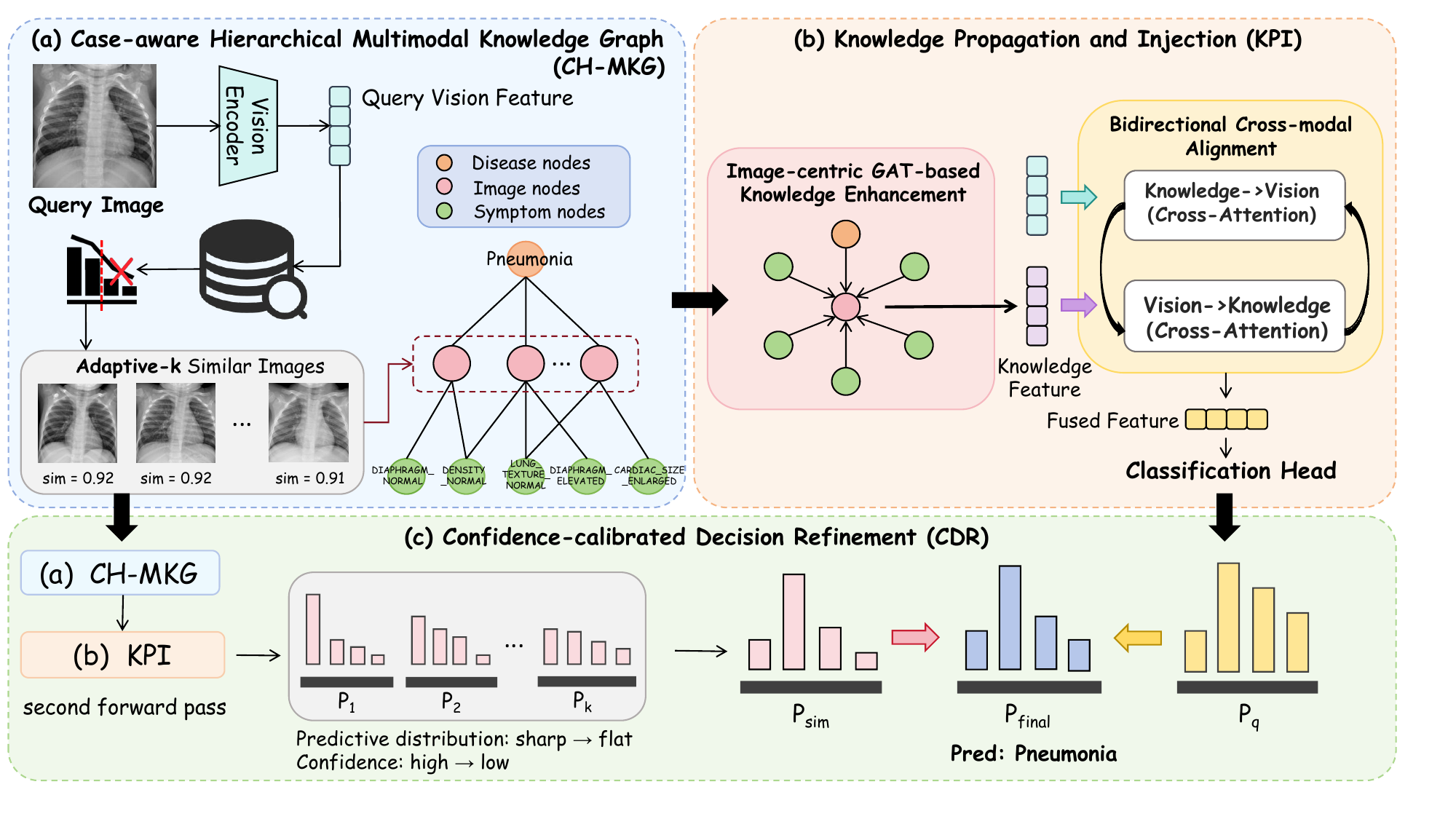} 
  
  \caption{The overall framework of MKG-CARE consists of three modules: (a) Case-aware Hierarchical Multimodal Knowledge Graph, (b) Knowledge Propagation and Injection, (c) Confidence-calibrated Decision Refinement.}
  \label{fig:overview}
\end{figure*}

\subsection{Overview}

As illustrated in Figure \ref{fig:overview}, our framework takes a query image as input and produces a final prediction $P_{\text{final}}$ through three stages. In the Case-aware Hierarchical Multimodal Knowledge Graph (CH-MKG) module, a vision encoder extracts visual features from the query image, which are used to adaptively retrieve relevant similar images. These retrieved images, along with their associated disease and symptom nodes, form a structured multimodal knowledge graph that organizes case-aware knowledge hierarchically. Next, in the Knowledge Propagation and Injection (KPI) module, an image-centric graph attention mechanism propagates information across image, symptom, and disease nodes to produce enriched knowledge features. These features are then aligned with the query visual feature via bidirectional cross-modal attention, and the resulting fused representation is fed to a classifier to produce an initial prediction $P_q$. Finally, in the Confidence-calibrated Decision Refinement (CDR) module, each retrieved similar case is passed through the same pipeline to obtain its prediction $P_i$. The module aggregates these predictions by considering both their similarity and their confidence, yielding an auxiliary distribution $P_{\mathrm{sim}}$. This is adaptively fused with $P_q$ to obtain the final $P_{\mathrm{final}}$. Thus, the model moves from isolated visual inference to case-aware reasoning with explicit utilization of historical similar cases and decision refinement.

\subsection{Case-aware Hierarchical Multimodal Knowledge Graph}

\subsubsection{Case-aware Knowledge Graph Construction.}
To simulate how clinicians reason by referring to prior experience with similar historical cases, we construct a case-aware hierarchical multimodal knowledge graph tailored to each dataset, serving as a clinician's memory bank. Specifically, we select representative image samples for each disease, which serve as image nodes in the graph. Each image node is then connected to its corresponding disease node and the associated symptom nodes derived from expert annotations. Edges encode two types of relations: image-disease associations indicating class membership, and image-symptom associations capturing observable clinical characteristics.

This forms a hierarchical multimodal knowledge graph organized from top to bottom: the top layer consists of disease nodes representing diagnostic classes, serving as semantic anchors; the middle layer consists of image nodes representing representative cases associated with each disease; and the bottom layer consists of symptom nodes representing fine-grained clinical attributes that describe visual patterns. This hierarchical design enables bidirectional information flow: disease nodes at the top propagate global semantic guidance downward, while symptom nodes at the bottom transmit fine-grained descriptive evidence upward. Such a multi-granularity organization enables structured reasoning by explicitly modeling the relationships among categories, cases, and clinical symptoms. Compared with text-only knowledge graphs, our formulation explicitly incorporates visual instances as entities, allowing tighter coupling between image features and medical knowledge.

\subsubsection{Adaptive Case Retrieval.}
Given a query image, we first encode it into a feature vector using a pre-trained MedMamba~\cite{yue2024medmamba} encoder. We then retrieve similar image nodes from a FAISS vector database, which is constructed offline using the representative image nodes extracted from the training set. Instead of using a fixed number of retrievals, we adopt an adaptive retrieval strategy~\cite{zhao2025mmkg} that determines the effective number of relevant cases per query. Specifically, let $S_i$ denote the similarity score of the $i$-th retrieved image node. We compute a similarity ratio as follows:
\begin{equation}
u_i = \log\left(\frac{S_i}{S_{i+1}}\right), i \in \{1, 2, \dots, k-1\}
\end{equation}
When $u_i$ exceeds a predefined threshold $u_{th}$, it indicates a sharp drop in similarity. We therefore truncate the retrieval list at position $i$, retaining only the most informative cases while filtering out noisy or weakly related samples.

After selecting the image nodes, we construct a query-specific subgraph by performing a one-hop expansion over the knowledge graph, incorporating connected symptom and disease nodes. This results in an informative image-centric subgraph that consists of the retrieved similar cases and their associated symptom and disease nodes, serving as a structured knowledge context for subsequent knowledge feature extraction.

This retrieval-and-expansion mechanism transforms the input from an isolated sample into a structured reasoning unit, thereby enabling the model to reason by referring to similar cases and leveraging their associated semantic knowledge. 

\subsection{Knowledge Propagation and Injection}
After constructing the retrieval subgraph, our goal is to effectively leverage it as a structured knowledge context to obtain concise knowledge representations of similar cases. To this end, we design a two-stage knowledge propagation and injection mechanism that first performs image-centric Graph Attention Network-based knowledge enhancement over the retrieval subgraph, and then conducts bidirectional cross-modal alignment between the resulting knowledge representation and the query visual features. This design enables cross-modal semantic enrichment and deep cross-case interaction, allowing visual evidence to be jointly interpreted with structured clinical knowledge.

\subsubsection{Image-centric Graph Attention Network-based Knowledge Enhancement}
We first perform message passing over the retrieval subgraph using a two-layer image-centric Graph Attention Network~\cite{velivckovic2017graph}. Let $\mathbf{h}_i^{(l)}$ denote the representation of node $i$ at layer $l$. The node features are initialized as follows: image nodes are initialized with MedMamba~\cite{yue2024medmamba} embeddings, while symptom and disease nodes are initialized using BioBERT~\cite{lee2020biobert}. The GAT layer updates node representations by aggregating information from their neighbors with attention weights:
\begin{equation}
\mathbf{h}_i^{(l+1)} = \sigma\left( \sum_{j \in \mathcal{N}(i)} \alpha_{ij}^{(l)} \mathbf{W}^{(l)} \mathbf{h}_j^{(l)} \right),
\end{equation}
where $\alpha_{ij}^{(l)}$ is the attention coefficient computed as:
\begin{equation}
\alpha_{ij}^{(l)} = \frac{\exp\left( \mathrm{LeakyReLU}\left( \mathbf{a}^{(l)\top} \left[ \mathbf{W}^{(l)}\mathbf{h}_i^{(l)} \mathbin{\big\|} \mathbf{W}^{(l)}\mathbf{h}_j^{(l)} \right] \right) \right)}{\sum_{k \in \mathcal{N}(i)} \exp(\cdot)}.
\end{equation}

Through two stacked GAT layers, information is propagated along the structure of the graph. This process enables retrieved image nodes to absorb medical semantics from symptom and disease nodes, while knowledge nodes simultaneously aggregate statistical evidence from multiple similar images, allowing the model to assign higher attention weights to symptoms that are more discriminative for classification, effectively amplifying their contribution while suppressing noisy symptom nodes. As a result, each retrieved image node is transformed into a knowledge-enhanced representation that integrates both visual similarity and semantic context.

To obtain a compact knowledge representation, we aggregate the enhanced representations of retrieved images into a single vector. Specifically, we compute similarity-aware weights based on the query embedding and perform normalized weighted aggregation:
\begin{equation}
\mathbf{z}_{kg} = \sum_{i=1}^K \tilde{w}_i \mathbf{h}_i^{(L)}, \quad \tilde{w}_i = \frac{\mathrm{sim}_i}{\sum_j \mathrm{sim}_j},
\end{equation}
where $\mathrm{sim}_i$ denotes the similarity between the query and the $i$-th retrieved image. This aggregated vector $\mathbf{z}_{kg}$ captures knowledge-aware evidence from multiple related cases, serving as a global knowledge representation for subsequent alignment.

\subsubsection{Bidirectional Cross-modal Alignment}

While the graph propagation stage aggregates textual semantics into the visual features of similar images to obtain knowledge-enhanced representations, it does not explicitly align the knowledge feature space with the query visual feature space. To bridge this gap, we design a two-layer bidirectional cross-modal attention module, where each layer consists of a CrossModalBlock with bidirectional interactions.

Given the query visual feature $\mathbf{v}_q \in \mathbb{R}^d$ and the knowledge representation $\mathbf{z}_{kg} \in \mathbb{R}^d$, each CrossModalBlock performs the following sequence of operations to enable interaction between them.

First, we update the knowledge representation conditioned on the query feature as context:
\begin{equation}
\mathbf{z}_{kg}^{(1)} = \mathrm{MHA}\big(\mathbf{Q} = \mathbf{z}_{kg},\ \mathbf{K} = \mathbf{v}_q,\ \mathbf{V} = \mathbf{v}_q\big),
\end{equation}
which allows the knowledge vector to incorporate visual evidence from the query image.

Next, we update the query representation using the newly updated knowledge feature:
\begin{equation}
\mathbf{v}_q^{(1)} = \mathrm{MHA}\big(\mathbf{Q} = \mathbf{v}_q,\ \mathbf{K} = \mathbf{z}_{kg}^{(1)},\ \mathbf{V} = \mathbf{z}_{kg}^{(1)}\big),
\end{equation}
enabling the visual representation to be explicitly guided by semantically enriched knowledge.

Updates are followed by residual connections and normalization. 
\begin{equation}
\mathbf{z}_{kg} = \mathrm{LayerNorm}\big(\mathbf{z}_{kg} + \mathbf{z}_{kg}^{(1)}\big), 
\mathbf{v}_q = \mathrm{LayerNorm}\big(\mathbf{v}_q + \mathbf{v}_q^{(1)}\big),
\end{equation}
and further refined by modality-specific MLP layers:
\begin{equation}
\mathbf{z}_{kg} = \mathrm{LayerNorm}\big(\mathbf{z}_{kg} + \mathrm{MLP}_{kg}(\mathbf{z}_{kg})\big),
\end{equation}
\begin{equation}
\mathbf{v}_q = \mathrm{LayerNorm}\big(\mathbf{v}_q + \mathrm{MLP}_{img}(\mathbf{v}_q)\big).
\end{equation}

We stack two such CrossModalBlocks to enable iterative interaction between modalities. This bidirectional design ensures that knowledge features are grounded in visual evidence, while visual features are semantically enriched by knowledge, leading to a well-aligned joint representation space.

Finally, we construct a fused representation by combining complementary information from both modalities. Specifically, we use four types of feature interactions:
\begin{equation}
\mathbf{h}_{fusion} = \mathrm{concat}\big(\mathbf{v}_q, \mathbf{z}_{kg}, |\mathbf{v}_q - \mathbf{z}_{kg}|, \mathbf{v}_q \odot \mathbf{z}_{kg}\big),
\end{equation}
where the absolute difference captures discrepancies, and element-wise product captures consistency between modalities.

The fused representation is then fed into a classification head:
\begin{equation}
\mathrm{logits}_q = \mathrm{MLP}\big(\mathbf{h}_{fusion}\big),
\end{equation}
which produces the initial prediction for the query image.

\subsection{Confidence-calibrated Decision Refinement}

Although retrieval provides additional evidence, the quality of retrieved samples varies, and noisy cases can mislead the model's diagnostic direction. Relying solely on similarity to the query is often insufficient, as highly similar cases may still be mislabeled or contain clinically irrelevant features, making similarity alone an unreliable indicator of sample reference value. Moreover, existing retrieval-augmented methods~\cite{tian2025ecg} lack mechanisms to quantify the contribution of each similar case to the final decision, making it difficult to interpret how the retrieved evidence influences the prediction and to understand the rationale behind the final decision.

To address these issues, we design a confidence-calibrated decision refinement mechanism that performs a second-pass inference over the retrieved cases, quantifying the reliability of each sample based on both its similarity to the query and its prediction confidence. These reliability weights are then used to fuse the retrieved predictions with the query result, thereby suppressing retrieval noise. Meanwhile, the confidence of similar cases provides an interpretable basis for understanding the final prediction, offering both robustness and interpretability.

We first obtain the initial prediction for the query image based on the query itself. Next, for each retrieved similar sample, we reuse its pre-constructed subgraph and perform the same forward inference pipeline to obtain its prediction distribution. The two prediction distributions are formulated as:
\begin{equation}
\mathbf{P}_q = \mathrm{softmax}(\mathrm{logits}_q),  \mathbf{P}_i = \mathrm{softmax}(\mathrm{logits}_i),
\end{equation}
where $\mathbf{P}_q$ denotes the baseline prediction of the query image and $\mathbf{P}_i$ denotes the prediction distribution of each retrieved sample.

To assess the reliability of each retrieved sample, we define its prediction confidence as:
\begin{equation}
c_i = \max(\mathbf{P}_i),
\end{equation}
which reflects the certainty of its prediction.

We compute the contribution weight of each retrieved sample by jointly considering its similarity to the query and its prediction confidence. Specifically, the weight is defined as:
\begin{equation}
w_i \propto (\mathrm{sim}_i)^\alpha \cdot (c_i)^\beta,
\end{equation}
where $\alpha$ and $\beta$ are hyperparameters that control the relative importance of similarity and confidence, respectively.

Based on these weights, we aggregate the predictions of retrieved samples into a unified distribution:
\begin{equation}
\mathbf{P}_{sim} = \frac{\sum_i w_i \mathbf{P}_i}{\sum_i w_i}.
\end{equation}

Finally, we combine the query prediction and the aggregated prediction from retrieved samples:
\begin{equation}
\mathbf{P}_{final} = (1-\lambda)\mathbf{P}_q + \lambda\mathbf{P}_{sim},
\end{equation}
and the final predicted label is obtained by $\arg\max(\mathbf{P}_{final})$.

This refinement mechanism improves robustness by emphasizing reliable and relevant retrieved samples, while suppressing noisy ones. Besides, the weighting scheme provides a quantitative measure of how each retrieved case contributes to the final decision, enhancing the interpretability in a case-based reasoning manner.

\section{Experiments}

\subsection{Experimental Setup}

\subsubsection{Datasets}
We evaluate our method on five widely-used medical image classification benchmarks: BreastMNIST~\cite{al2020dataset}, DermaMNIST~\cite{yang2023medmnist}, Kvasir~\cite{pogorelov2017kvasir}, PAD\_UFES\_20~\cite{pacheco2020pad}, and RetinaMNIST~\cite{yang2023medmnist}, spanning dermatology, ophthalmology, pathology, and gastrointestinal imaging. These datasets vary in class granularity, data distribution, and visual complexity, exhibiting challenges such as class imbalance, fine-grained visual differences, and high intra-class variation, which together provide a comprehensive testbed for evaluating generalization, robustness, and knowledge-enhanced reasoning. We summarize the five datasets and the corresponding knowledge graph scales (number of entities and edges) in Table~\ref{tab:datasets}.

\begin{table}[htbp]
  \centering
  \caption{Summary of five medical image classification datasets used in our experiments.}
  \label{tab:datasets}
  \resizebox{\linewidth}{!}{
  \begin{tabular}{lcccccc}
    \toprule
    Dataset & Domain & Classes & Size & KG Entities & KG Edges \\
    \midrule
    BreastMNIST & Pathology & 2 & 780 & 60 & 271 \\
    DermaMNIST & Dermatology & 7 & 10,015 & 163 & 1088 \\
    Kvasir & Gastrointestinal & 8 & 8,000 & 162 & 984 \\
    PAD\_UFES\_20 & Dermatology & 6 & 2,298 & 153 & 1000 \\
    RetinaMNIST & Ophthalmology & 5 & 1,600 & 106 & 1040 \\
    \bottomrule
  \end{tabular}
  }
\end{table}

\subsubsection{Implementation Details}
Our experiments are conducted using PyTorch on an NVIDIA RTX A800 GPU. For fair comparison, all input images are resized to $224 \times 224 \times 3$. For efficient retrieval, representative image features are indexed using FAISS.

The knowledge-aware feature propagation module uses a two-layer Graph Attention Network with multi-head attention. Specifically, each layer employs 4 attention heads, and the hidden dimension is fixed at 768. A dropout rate of 0.2 is applied within the attention mechanism to mitigate overfitting. For cross-modal interaction, we employ a two-layer bidirectional CrossModalBlock. Each layer consists of multi-head cross-attention with 4 heads, an embedding dimension of 768, and an attention dropout rate of 0.2. This is followed by residual connections, LayerNorm, and modality-specific MLPs. Each MLP is implemented as a two-layer feed-forward network with ReLU activation. A dropout rate of 0.4 is applied for further refinement within each modality.

After cross-modal alignment, we construct a fused representation by combining the query visual feature and knowledge feature, resulting in a $4 \times 768$-dimensional vector. This fused feature is then fed into a two-layer MLP classifier with ReLU activation and dropout (0.4) to produce the final logits, and the model is trained using cross-entropy loss.

For optimization, we use the AdamW optimizer with an initial learning rate of $5 \times 10^{-5}$ and weight decay of 0.02. A StepLR scheduler is applied with step size 25 and decay factor 0.5 to ensure stable convergence. The model is trained with a batch size of 64 for up to 40 epochs, with early stopping based on validation performance to prevent overfitting. Gradient clipping is applied to stabilize optimization. For the confidence-calibrated refinement module, the weighting hyperparameters are set to $\alpha = 1$ and $\beta = 1$ by default.

\subsubsection{Baselines}
To evaluate the proposed method, we compare it with diverse baselines spanning multiple paradigms, including conventional vision-based architectures, large multimodal models (both general-purpose and medical-specific), and knowledge-enhanced approaches. Specifically, we include several representative \textbf{vision-based deep learning models}, such as Swin-Transformer~\cite{liu2021swin}, EfficientNetV2~\cite{tan2021efficientnetv2}, ConvNeXt~\cite{liu2022convnet}, and VMamba~\cite{liu2024vmamba}, which are widely used for medical image classification. We also report the performance of MedMamba~\cite{yue2024medmamba} as a backbone baseline to verify the improvement brought by our knowledge-enhanced framework. For \textbf{large multimodal models}, we evaluate both general-purpose baselines including GPT-5.1~\cite{openai_gpt_5}, Gemini-3-pro~\cite{google2025gemini3}, and Qwen3.5-plus~\cite{qwen_team_qwen35plus}, as well as medical-specific models such as HealthGPT~\cite{lin2025healthgpt}, HuatuoGPT~\cite{zhang2023huatuogpt}, and MedGemma~\cite{sellergren2025medgemma}, which are designed for healthcare applications. These models leverage large-scale multimodal pretraining and demonstrate strong generalization capability. Furthermore, we compare with a recent \textbf{knowledge-enhanced method} KEM~\cite{zhang2025knowledge} that incorporates textual knowledge generated by large language models into visual classification, serving as a reference for evaluating structured knowledge integration. This diverse set of baselines ensures a fair and comprehensive comparison across different modeling paradigms.

\subsubsection{Evaluation Metrics}
We evaluate all methods using two widely adopted metrics for medical image classification: Overall Accuracy (OA) and Macro-AUC. OA measures the proportion of correctly classified samples, reflecting general prediction performance. However, in medical datasets with imbalanced class distributions, OA alone may not fully capture the model’s ability to distinguish different disease categories. Therefore, we additionally report Macro-AUC, which computes the average area under the ROC curve across all classes, treating each class equally. This metric captures per-class discriminative capability, providing a more comprehensive evaluation especially for rare or underrepresented classes. Together, these two metrics offer a balanced assessment of both overall performance and class-wise robustness.

\subsection{Main Results}

\begin{table*}[htbp]
\centering
\small
\renewcommand{\arraystretch}{0.95}
\caption{Performance comparison on five medical image classification benchmarks. All results are reported as OA and AUC, where AUC denotes Macro-AUC for multi-class datasets. The best results are highlighted in bold, and the second-best results are underlined. Rows shaded in \colorbox[HTML]{DCDCDC}{gray} and \colorbox[HTML]{D7F6FF}{blue} indicate the knowledge-enhanced method and ours, respectively.}
\label{tab:main_result}
\resizebox{\linewidth}{!}{
\begin{tabular}{lcccccccccc}
\toprule
\textbf{Model}
& \multicolumn{2}{c}{\textbf{BreastMNIST}} & \multicolumn{2}{c}{\textbf{DermaMNIST}} & \multicolumn{2}{c}{\textbf{Kvasir}} & \multicolumn{2}{c}{\textbf{PAD\_UFES\_20}} & \multicolumn{2}{c}{\textbf{RetinaMNIST}} \\
\cmidrule(lr){2-3} \cmidrule(lr){4-5} \cmidrule(lr){6-7} \cmidrule(lr){8-9} \cmidrule(lr){10-11}
& \textbf{OA} & \textbf{AUC} & \textbf{OA} & \textbf{AUC} & \textbf{OA} & \textbf{AUC} & \textbf{OA} & \textbf{AUC} & \textbf{OA} & \textbf{AUC} \\
\midrule

\textbf{Vision-based} & & & & & & & & & & \\
Swin-Transformer (2021)         & 75.6 & 78.8 & 77.8 & 93.8 & 75.2 & 97.0 & 59.7 & 82.5 & 53.5 & 72.5 \\
EfficientNetV2 (2021)           & 84.0 & 76.4 & 75.1 & 90.1 & 76.3 & 96.3 & 51.7 & 74.1 & 51.3 & 70.5 \\
ConvNeXt (2022)                 & 83.3 & 81.7 & 67.0 & 91.9 & 72.5 & 96.6 & 49.4 & 77.7 & 51.8 & 69.8 \\
 VMamba (2024)                   & 82.1 & 80.4 & 70.3 & 89.0 & 69.8 & 96.2 & 52.3 & 82.6 & 52.3 & 73.2 \\
 MedMamba (2024)                 & 85.3 & 82.5 & 77.9 & 91.7 & 79.3 & \underline{97.6} & 58.8 & 80.8 & 54.3 & 74.7 \\

\midrule

\textbf{Med-LVLM} & & & & & & & & & & \\
Health-GPT-XL32 (2024)          & 37.2 & 56.3 & 33.3 & 52.1 & 34.5 & 62.6 & 39.2 & 61.9 & 40.3 & 49.9 \\
HuatuoGPT-Vision-34B (2024)     & 35.3 & 55.0 & 61.7 & 65.0 & 47.3 & 69.9 & 54.9 & 65.3 & 36.5 & 60.0 \\
medgemma-1.5-4b-it (2026)       & 73.7 & 51.2 & 54.9 & 60.6 & 29.2 & 59.8 & 53.5 & 69.5 & 51.0 & 62.3 \\

\midrule

\textbf{General LVLM} & & & & & & & & & & \\
GPT-5.1 (2025)                  & 64.7 & 70.9 & 72.7 & 84.3 & 80.3 & 85.2 & 57.7 & 74.2 & 28.5 & 48.3 \\
Gemini-3-pro (2025)                 & 80.1 & 77.6 & 75.6 & 85.9 & 81.6 & 87.1 & 61.1 & 70.7 & 31.5 & 58.0 \\
Qwen3.5-plus (2026)             & 75.0 & 72.4 & 56.4 & 71.9 & 63.4 & 79.1 & 59.1 & 72.4 & 47.0 & 55.7 \\

\midrule

\textbf{Knowledge-enhanced} & & & & & & & & & & \\
\rowcolor[HTML]{DCDCDC} KEM (2025)                      & \underline{86.5} & \textbf{90.9} & \underline{81.7} & \underline{95.8} & \underline{82.3} & 97.5 & \underline{62.0} & \underline{82.6} & \underline{55.8} & \underline{75.6} \\

\midrule

\textbf{Ours} & & & & & & & & & & \\
\rowcolor[HTML]{D7F6FF} MKG-CARE                        & \textbf{87.2} & \underline{83.1} & \textbf{83.6} & \textbf{96.1} & \textbf{84.0} & \textbf{98.4} & \textbf{63.1} & \textbf{83.3} & \textbf{58.8} & \textbf{76.0} \\

\bottomrule
\end{tabular}
}
\end{table*}

\subsubsection{Overall Performance}

Table \ref{tab:main_result} reports the comparison results of all methods on five medical image classification benchmarks. We evaluate all methods using Overall Accuracy (OA) and AUC, comparing our approach against a comprehensive set of baselines, including vision-based models, medical/general Large Vision-Language Models (LVLMs), and knowledge-enhanced methods.

Overall, our method achieves the best OA performance on all five datasets and the best AUC on four of them, demonstrating consistent performance gains across diverse medical domains and classification settings. These results validate the effectiveness of case-aware reasoning in medical image classification.

Notably, while achieving the highest OA on BreastMNIST, our method yields the second-best Macro-AUC. This is attributed to the dataset's artificially merged binary structure. Specifically, our model's fine-grained semantic disentanglement inadvertently separates samples within the heterogeneous negative class (merged from normal and benign), which slightly affects ranking-based metric AUC without compromising the final classification decisions.

\subsubsection{Comparison with Vision-based Models}


Compared to conventional vision architectures (e.g., EfficientNetV2, Swin-Transformer, ConvNeXt, and VMamba), our method achieves consistent improvements. These baselines rely purely on visual features and often struggle with the high intra-class variation common in medical imaging. More importantly, our approach significantly outperforms MedMamba~\cite{yue2024medmamba}, which serves as our visual backbone. This demonstrates that our performance gains are not merely derived from a stronger encoder, but primarily from the effective integration of structured knowledge and similar-case reasoning.

\subsubsection{Comparison with LVLMs}

We further compare with medical-specific (Health-GPT, MedGemma, HuatuoGPT) and general-purpose (GPT-5.1, Gemini-3, Qwen3.5-plus) LVLMs. While these billion-parameter models possess strong general capabilities, they often lack stability in specialized medical scenarios requiring fine-grained visual perception. Furthermore, they incur substantial computational cost. In contrast, our method maintains a computationally efficient architecture. By explicitly grounding predictions in retrieved cases and structured knowledge, our approach provides more reliable and domain-aware representations, outperforming large-scale LVLMs while remaining computationally efficient.



\subsubsection{Comparison with Knowledge-enhanced Method}

Compared to the knowledge-enhanced baseline KEM~\cite{zhang2025knowledge}, our method exhibits consistent superiority. KEM processes images in isolation and relies on LLM-generated symptom descriptions, which lack structural hierarchy and are prone to hallucinations. Our framework overcomes these limitations through three key designs: (1) we retrieve and reason over historical similar cases rather than performing isolated inference; (2) we utilize expert-annotated medical knowledge, eliminating hallucination risks; and (3) we construct a structured graph with an attention mechanism, enabling the model to dynamically prioritize informative symptoms over noisy ones. Collectively, these advantages lead to our leading performance.

\subsection{Ablation Study}

\begin{table}[htbp]
  \centering
  \caption{\textbf{Ablation study of key components.} We report overall accuracy (OA) on the Kvasir and RetinaMNIST datasets. CH-MKG, KPI, and CDR denote the three modules introduced in our method.}
  \label{tab:ablation}
  \begin{tabular}{lcc}
    \toprule
    Models & Kvasir & RetinaMNIST \\
    \cmidrule(lr){2-2} \cmidrule(lr){3-3}
    & OA & OA \\
    \midrule
    Full Model & \textbf{84.0} & \textbf{58.8} \\
    w/o. CDR & 83.6 & 58.0 \\
    w/o. KPI \& CDR & 82.7 & 56.3 \\
    w/o. CH-MKG \& KPI \& CDR & 82.0 & 56.0 \\
    \bottomrule
  \end{tabular}
\end{table}

We conduct ablation studies on the Kvasir and RetinaMNIST datasets to systematically evaluate the contribution of each core module in our framework: the Case-aware Hierarchical Multimodal Knowledge Graph (CH-MKG), Knowledge Propagation and Injection (KPI), and Confidence-calibrated Decision Refinement (CDR). As shown in Table~\ref{tab:ablation}, we adopt a \emph{progressive removal} strategy, starting from the full model and successively disabling individual modules to analyze their impact on overall accuracy (OA).

\subsubsection{Impact of the CDR Module} 

Removing CDR leads to an OA drop on both datasets (Kvasir: 84.0 -> 83.6; RetinaMNIST: 58.8 -> 58.0). This degradation highlights CDR's role in enhancing prediction reliability. By explicitly weighting retrieved cases based on a combination of similarity and confidence, it effectively mitigates the influence of noisy or misleading samples. This demonstrates that confidence-aware case aggregation is essential for robust case-based reasoning, especially when retrieval quality varies.

\subsubsection{Impact of the KPI Module} 

Further disabling KPI causes a more pronounced decline (Kvasir: 83.6 -> 82.7; RetinaMNIST: 58.0 -> 56.3). Without KPI, the model degenerates into naive similarity-weighted aggregation and shallow feature concatenation. This indicates that without proper knowledge propagation, the model fails to leverage case-level semantic knowledge and instead relies on superficial feature matching. This demonstrates that KPI is essential for deep cross-modal alignment and propagating textual medical semantics into visual representations.

\subsubsection{Impact of the CH-MKG Module} 
Next, we further disable the CH-MKG module (w/o. CH-MKG \& KPI \& CDR), where a portion of informative nodes (including representative image nodes and associated symptom nodes) is removed from the knowledge graph. This results in additional performance decline (Kvasir: 82.7 $\rightarrow$ 82.0; RetinaMNIST: 56.3 $\rightarrow$ 56.0), indicating that the effectiveness of the framework is tightly coupled with the availability of high-quality, semantically meaningful nodes. By removing these nodes, the retrieved subgraph becomes less informative, weakening both the case-level evidence provided by similar images and the complementary clinical semantics introduced by symptom nodes.  This observation confirms that our framework does not merely leverage the graph structure, but fundamentally relies on the semantic content encoded in carefully constructed multimodal nodes.


\subsection{Case Study}
\subsubsection{Visualization of Confidence Calibration with Knowledge Graph Reasoning}

\begin{figure*}[t]
  \centering
  \includegraphics[
    width=0.92\textwidth,
    height=11cm,
    keepaspectratio
  ]{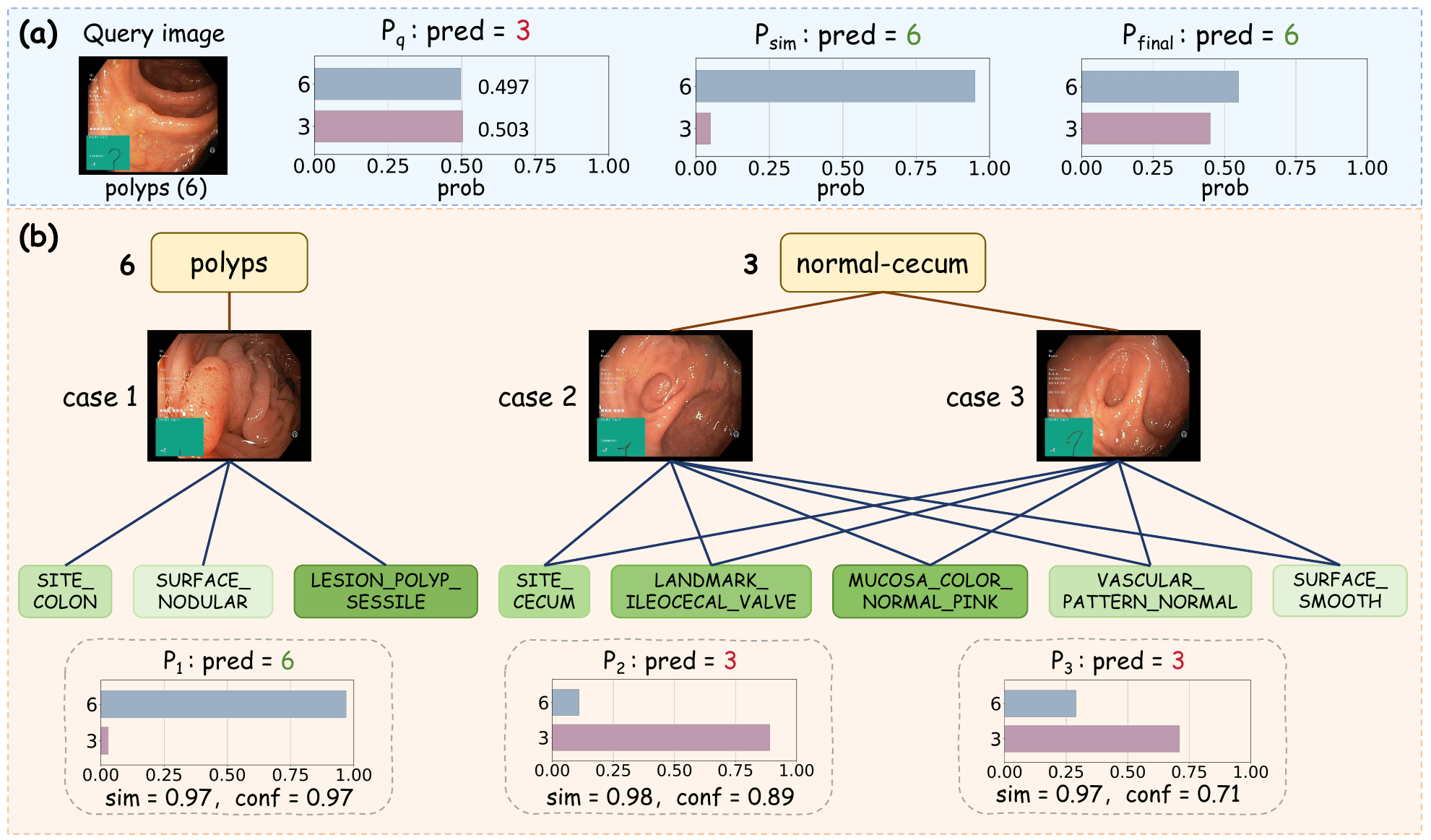}
  
  \caption{\textbf{Visualization of confidence calibration with knowledge graph reasoning on the Kvasir dataset.} (a) Prediction evolution of the query image from the initial prediction $P_q$ to the retrieved-case aggregated prediction $P_{sim}$ and the final refined prediction $P_{final}$. (b) Retrieved case-aware hierarchical multimodal knowledge graph, showing the relationships among disease nodes, retrieved image nodes, and symptom nodes. The color intensity of symptom nodes reflects their learned importance during knowledge propagation. Each retrieved case is annotated with its individual prediction distribution, similarity score and prediction confidence, illustrating how CDR module estimates the reliability of retrieved samples during decision refinement.}
  \label{fig:interpret_case}
\end{figure*}

Figure~\ref{fig:interpret_case} illustrates the interpretable reasoning process of MKG-CARE using a representative Kvasir case. As shown in Figure~\ref{fig:interpret_case} (a), the initial prediction $P_q$ incorrectly favors class 3 over the ground-truth class 6. After incorporating evidence from the retrieved cases, the aggregated prediction $P_{\text{sim}}$ shifts toward class 6, leading to the correct final prediction $P_{\text{final}}$. Figure~\ref{fig:interpret_case}(b) further visualizes how this refinement is achieved within the case-aware knowledge graph. Retrieved image nodes are connected to their associated disease and symptom nodes, where the intensity of symptom nodes reflects their learned importance during knowledge propagation. Although multiple retrieved cases exhibit similarly high visual similarity to the query, their diagnostic reliability differs substantially. The CDR module therefore jointly considers similarity and prediction confidence when determining their contributions. In particular, Case 1, which exhibits both high similarity and high confidence for class 6, provides the strongest support for correcting the query prediction, whereas cases with conflicting or uncertain predictions are down-weighted despite their high similarity. This visualization demonstrates that MKG-CARE can distinguish useful evidence from visually similar but less reliable cases, thereby suppressing retrieval noise while providing interpretable case-level support for the final diagnosis.

\subsubsection{Lesion-aware Attention Visualization}

\begin{figure*}[t]  
    \centering
    \includegraphics[width=0.8\textwidth, height=11.5cm, keepaspectratio]{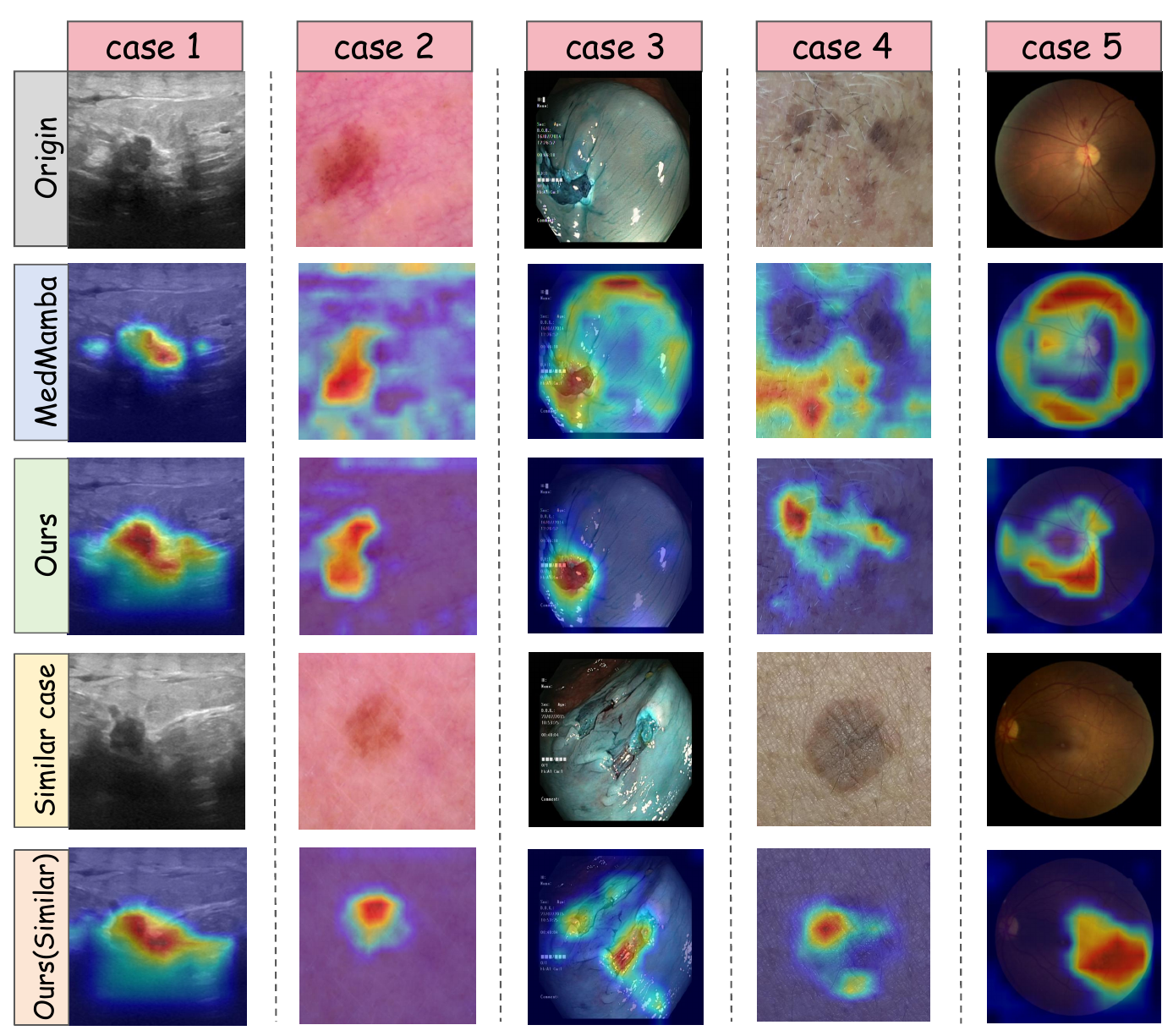}
    
    \caption{Grad-CAM visualization across five datasets (Case 1–5 correspond to BreastMNIST, DermaMNIST, Kvasir, PAD\_UFES\_20, and RetinaMNIST, respectively). From top to bottom: original image, MedMamba attention map, our method’s attention map, most similar retrieved image, and our method’s attention map on the retrieved image.}
    \label{fig:attention_visualization}
\end{figure*}

Figure \ref{fig:attention_visualization} uses Grad-CAM to compare the visual reasoning of our method against the MedMamba baseline across five datasets. Consistently, MedMamba produces diffuse attention maps, often highlighting irrelevant background areas or completely missing subtle pathological cues (e.g., Case 4). By contrast, our retrieval-augmented framework generates highly concentrated and semantically accurate activations. It precisely localizes clinically critical regions across diverse domains, such as the hypoechoic mass with irregular margins in ultrasound (Case 1), heterogeneous pigmentation in dermoscopy (Cases 2 and 4), and specific pathological structures in endoscopy and fundus imaging (Cases 3 and 5). Furthermore, the attention patterns on the query images closely align with those on the retrieved similar cases. This provides strong qualitative evidence that our model effectively transfers cross-case visual knowledge, utilizing analogous disease patterns to refine its lesion focus.

\subsection{Further Analysis}

\begin{figure}[t]  
    \centering
    \includegraphics[width=1.0\linewidth, height=12.5cm, keepaspectratio]{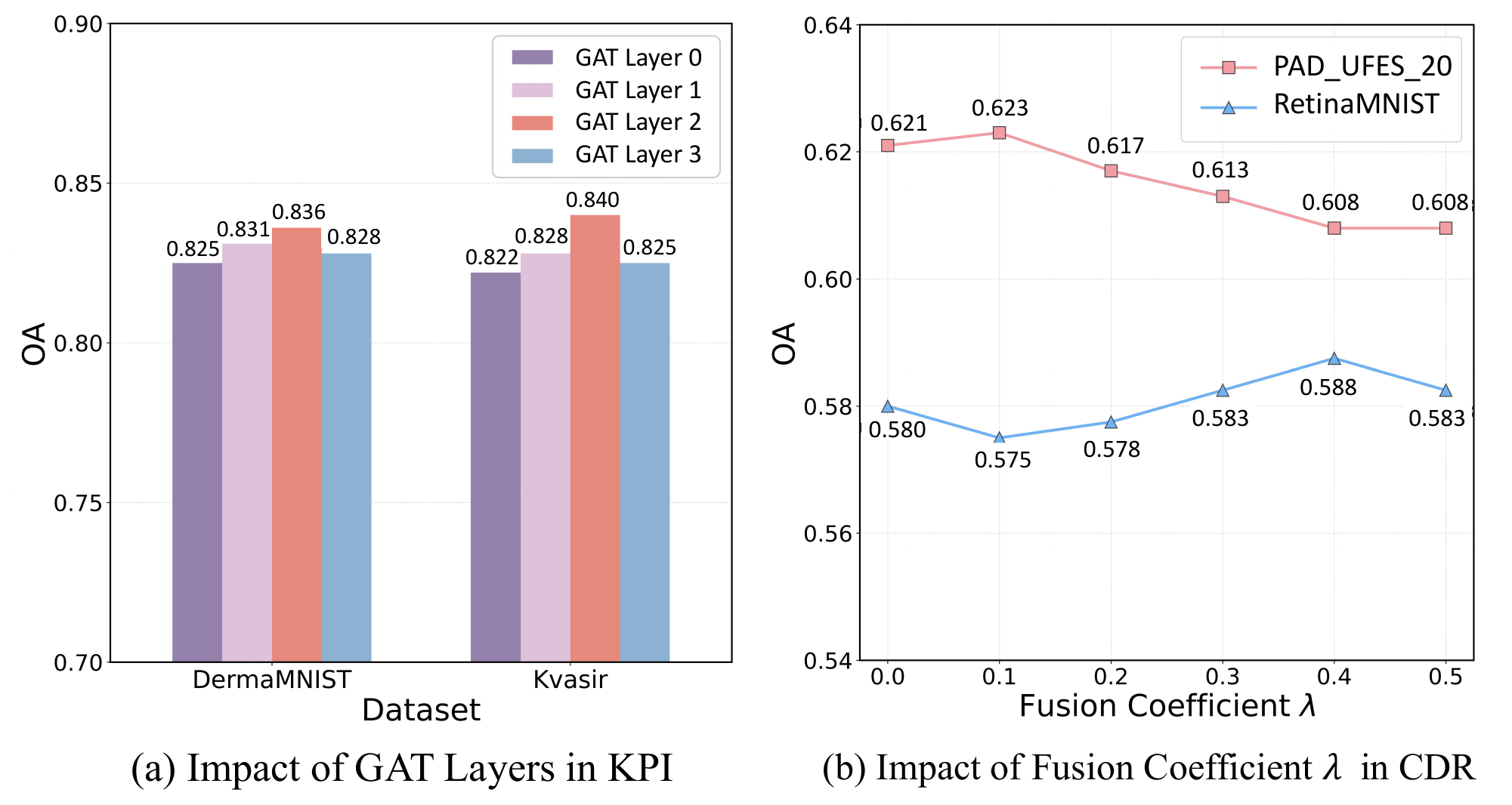}
    \caption{Parameter Sensitivity Analysis of MKG-CARE.}
    \label{fig:further_analysis}
\end{figure}

\subsubsection{Impact of GAT Layers in KPI}

We further analyze the influence of the number of GAT layers in the KPI module, with results illustrated in Figure \ref{fig:further_analysis}(a).

With 0 layers, the model degenerates into a visual-only retrieval baseline without textual knowledge, yielding the lowest performance (0.825 on DermaMNIST and 0.822 on Kvasir). This indicates that relying solely on visual similarity is insufficient for capturing clinically meaningful semantics. Performance improves with 1 layer and peaks at 2 layers (0.836 and 0.840, respectively). Unlike single-layer propagation where knowledge nodes lack cross-case aggregation, the 2-layer GAT optimally facilitates bidirectional message passing: knowledge nodes first aggregate statistical evidence from multiple images, and then propagate enriched, cross-case semantics back to the image representations. However, further increasing the depth to 3 layers causes a slight performance drop (0.828 / 0.825), primarily due to the well-known over-smoothing effect in graph networks, which weakens node distinctiveness. Therefore, we adopt a 2-layer architecture to balance effective knowledge aggregation and noise control.

\subsubsection{Impact of Fusion Coefficient $\lambda$ in CDR}

We investigate the effect of the fusion coefficient $\lambda$, which controls the contribution of retrieved-case predictions $P_{\text{sim}}$ to the final prediction. As shown in Figure~\ref{fig:further_analysis} (b), PAD\_UFES\_20 achieves the best performance at a small $\lambda=0.1$, suggesting that the query image itself provides sufficiently reliable evidence and excessive retrieval information may introduce noise. In contrast, RetinaMNIST benefits from a larger $\lambda$, with performance peaking at $\lambda=0.4$, as its subtle inter-class differences make retrieved cases a useful complement to the query. However, further increasing $\lambda$ degrades performance, indicating that over-reliance on retrieved cases can amplify unreliable evidence. Overall, $\lambda$ balances query evidence and retrieval-based reasoning across different diagnostic scenarios.

\section{Conclusion}


In this paper, we propose MKG-CARE, a case-aware framework that leverages multimodal knowledge graphs for explainable medical image diagnosis. MKG-CARE organizes diseases, images, and symptoms in a hierarchical case-aware graph, propagates and aligns structured knowledge with query visual features, and refines predictions by calibrating the reliability of retrieved cases. Extensive experiments across multiple medical image datasets demonstrate consistent performance improvements, while qualitative analyses further show interpretable case-level evidence attribution and clinically meaningful attention. These results highlight the value of integrating structured multimodal knowledge with case-based reasoning for more reliable and interpretable medical image diagnosis. Future work will extend MKG-CARE to more complex 3D imaging modalities, such as CT and MRI.

\begin{acks}
The research was partially supported by Anhui Provincial Science and Technology Innovation Program 202523o09050019 and 202523o09050015,  and the China National Natural Science Foundation with no. 92567301 and 62132018. We also gratefully acknowledge support from the National Center for Interdisciplinary Studies and Talents Cultivation, USTC.
\end{acks}

\section*{GenAI Usage Disclosure}
During the preparation of this work, the authors used ChatGPT\,\textsuperscript{1} to polish the paper and subsequently reviewed and edited the content. They take full responsibility for the final publication.

\footnotetext[1]{https://chat.openai.com}

\bibliographystyle{ACM-Reference-Format}
\bibliography{ref}










\end{document}